\documentclass[11pt]{article}
\usepackage[margin=1in]{geometry}
\usepackage{amsmath,amssymb,graphicx,hyperref,xcolor,booktabs}
\usepackage{algorithm}
\usepackage{algorithmic}
\usepackage{multirow}
\usepackage{xcolor}
\usepackage{subcaption}

\title{FSC-Net: Fast-Slow Consolidation Networks \\for Continual Learning}

\author{
  Mohamed El Gorrim \\
  \texttt{elgorrim.mohamed@etu.uae.ac.ma} \\
}

\begin{document}

\maketitle

\begin{abstract}
Continual learning remains challenging due to catastrophic forgetting, where neural networks lose previously acquired knowledge when learning new tasks. Inspired by memory consolidation in neuroscience, we propose \textbf{FSC-Net} (Fast-Slow Consolidation Networks), a dual-network architecture that separates rapid task learning from gradual knowledge consolidation. Our method employs a fast network (NN1) for immediate adaptation to new tasks and a slow network (NN2) that consolidates knowledge through distillation and replay. Within the family of MLP-based NN1 variants we evaluated, consolidation effectiveness is driven more by methodology than architectural embellishments---a simple MLP outperforms more complex similarity-gated variants by 1.2pp. Through systematic hyperparameter analysis, we observed empirically that \emph{pure replay without distillation} during consolidation achieves superior performance, consistent with the hypothesis that distillation from the fast network introduces recency bias. On Split-MNIST (30 seeds), FSC-Net achieves \textbf{91.71\% ± 0.62\%} retention accuracy, a +4.27pp gain over the fast network alone (87.43\% ± 1.27\%, paired $t=23.585$, $p < 1\times10^{-10}$). On Split-CIFAR-10 (5 seeds), our method achieves \textbf{33.31\% ± 0.38\%} retention with an +8.20pp gain over the fast network alone (25.11\% ± 1.61\%, paired $t=9.75$, $p < 10^{-3}$), demonstrating +8.20pp gain, though absolute performance (33.31\%) remains modest and below random expectation, highlighting need for stronger backbones. Our results provide empirical evidence that the dual-timescale consolidation mechanism, rather than architectural complexity, is central to mitigating catastrophic forgetting in this setting.
\end{abstract}

\section{Introduction}

Neural networks suffer from \emph{catastrophic forgetting} when trained on sequential tasks~\cite{mccloskey1989catastrophic,french1999catastrophic}. As new tasks are learned, previously acquired knowledge deteriorates rapidly, limiting the deployment of deep learning systems in dynamic, non-stationary environments. This challenge has motivated extensive research in continual learning~\cite{parisi2019continual,delange2021continual}.

Biological brains avoid catastrophic forgetting through \emph{memory consolidation}---the gradual transfer of information from fast-learning hippocampal systems to slow-learning neocortical structures during sleep~\cite{squire2015memory,mcclelland1995there}. This dual-timescale learning enables rapid adaptation to new experiences while preserving long-term knowledge stability.

Existing continual learning methods employ various strategies: regularization-based approaches~\cite{kirkpatrick2017overcoming,zenke2017continual} constrain weight updates to preserve task-specific parameters; replay-based methods~\cite{rolnick2019experience,chaudhry2019tiny} store and rehearse past examples; dynamic architectures~\cite{rusu2016progressive,yoon2017lifelong} expand network capacity for new tasks. However, these methods often require task-specific components or complex architectural modifications.

\textbf{Our Contribution.} We propose FSC-Net, a simple yet effective dual-network framework inspired by memory consolidation. Our key contributions are:

\begin{enumerate}
    \item \textbf{Consolidation Across Simple Architectures:} Within the family of MLP-based NN1 variants we evaluate, consolidation effectiveness is driven more by methodology than architectural embellishments. A lightweight MLP outperforms similarity-gated variants, indicating that the dual-timescale protocol is effective without specialised layers.
    
    \item \textbf{Dual-Timescale Learning:} Fast network (NN1) rapidly adapts to new tasks while slow network (NN2) consolidates knowledge through distillation and replay, achieving +4.27\% improvement (paired t-test: $t=23.585$, $p < 1\times10^{-10}$) on Split-MNIST (30 seeds).
    
    \item \textbf{Consolidation Without Distillation:} Through systematic hyperparameter analysis, we observed that pure replay ($\lambda=0$) during consolidation outperforms distillation-augmented replay ($\lambda=0.5$) by +1.26\% on MNIST and +1.76\% on CIFAR-10. We hypothesise that distillation from the fast network introduces recency bias, and we report empirical evidence supporting this behaviour.
    
    \item \textbf{Robust Empirical Validation:} We provide statistical rigor through a 30-seed evaluation on Split-MNIST (91.71\% ± 0.62\% retention) and 5-seed validation on CIFAR-10 (33.31\% ± 0.38\%). While CIFAR-10 absolute performance remains modest, the consistent +8.20pp improvement and tight variance (±0.38\%) demonstrate the stability of our consolidation mechanism.
    
    \item \textbf{Simplicity and Practicality:} Our method uses standard components (MLP, knowledge distillation, replay buffer) without requiring task boundaries, making it practical for real-world deployment.
\end{enumerate}

Our work reframes continual learning as a consolidation problem rather than an architecture design problem, opening avenues for applying this methodology to diverse network architectures.

\section{Related Work}

\subsection{Continual Learning Approaches}

\textbf{Regularization-based methods} constrain weight updates to preserve task-specific knowledge. Elastic Weight Consolidation (EWC)~\cite{kirkpatrick2017overcoming} penalizes changes to important parameters identified through Fisher information. Synaptic Intelligence~\cite{zenke2017continual} tracks parameter importance online. PackNet~\cite{mallya2018packnet} prunes and freezes network weights. While effective, these methods require task boundaries and struggle with task similarity.

\textbf{Replay-based methods} store representative samples from previous tasks. Experience Replay~\cite{rolnick2019experience} maintains a memory buffer of past examples. Generative Replay~\cite{shin2017continual} uses generative models to synthesize past data. Our approach combines replay with knowledge distillation for enhanced consolidation.

\textbf{Dynamic architectures} allocate dedicated capacity for each task. Progressive Neural Networks~\cite{rusu2016progressive} add new columns per task. DEN~\cite{yoon2017lifelong} dynamically expands network capacity. These methods avoid forgetting through task isolation but suffer from unbounded growth.

\textbf{Meta-learning approaches} learn to learn continually. Meta-Experience Replay~\cite{riemer2018learning} meta-learns from task distribution. OML~\cite{javed2019meta} meta-trains representations for fast adaptation. These require meta-training data unavailable in many scenarios.

\subsection{Neuroscience-Inspired Methods}

Our work is inspired by \textbf{complementary learning systems theory}~\cite{mcclelland1995there,kumaran2016learning}, which posits that the mammalian brain employs dual learning systems: a fast hippocampal system for rapid encoding and a slow neocortical system for gradual consolidation during sleep.

Several works have explored this concept computationally. Ans \& Rousset~\cite{ans2000neural} proposed a dual-network model with pseudorehearsal. Robins~\cite{robins1995catastrophic} demonstrated consolidation through interleaved learning. More recently, Kemker \& Kanan~\cite{kemker2018fearnet} proposed FearNet with separate short-term and long-term memory systems.

Our approach differs by: (1) demonstrating architecture-agnostic effectiveness through ablation studies, (2) employing knowledge distillation for consolidation, and (3) providing rigorous statistical validation across multiple seeds.

\section{Method: FSC-Net}

\subsection{Overview}

\begin{figure}[t]
\centering
\includegraphics[width=1\linewidth]{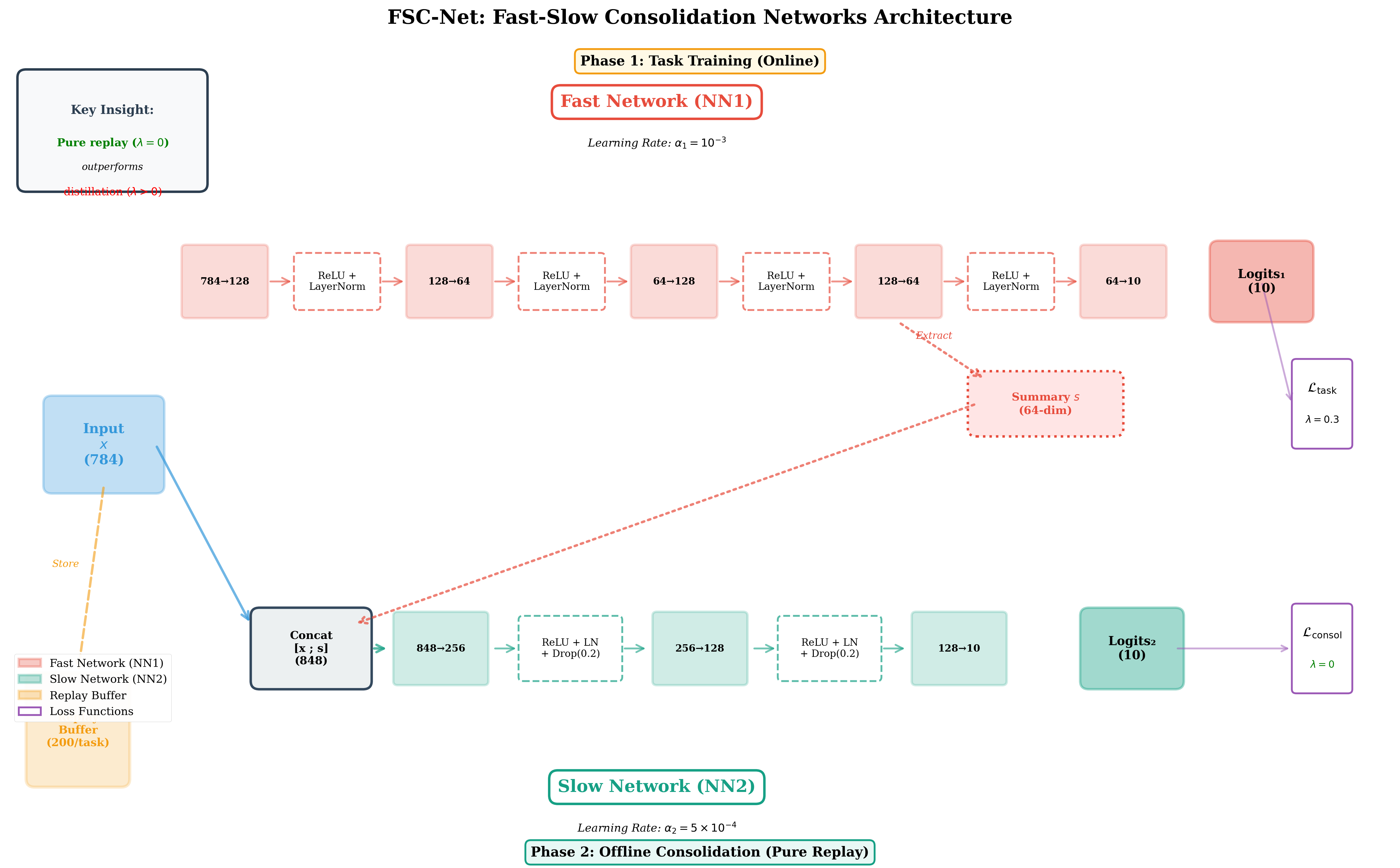}
\caption{FSC-Net architecture overview. The system employs dual networks: NN1 (fast, red) rapidly adapts to new tasks with a high learning rate ($10^{-3}$), while NN2 (slow, teal) consolidates knowledge with a lower learning rate ($5\times10^{-4}$). Input $x$ feeds both networks, with NN1 providing a summary embedding $s$ to NN2. The replay buffer stores samples from all tasks for offline consolidation. During task training, knowledge distillation ($\lambda=0.3$) helps NN2 track NN1's adaptation. During offline consolidation, pure replay ($\lambda=0$) provides superior performance by avoiding recency bias from NN1's task-specific predictions.}
\label{fig:architecture}
\end{figure}

FSC-Net consists of two networks operating at different timescales (Figure~\ref{fig:architecture}):

\begin{itemize}
    \item \textbf{Fast Network (NN1):} Rapidly adapts to new tasks with aggressive learning rates and frequent updates. Uses standard MLP architecture with 64-dimensional embeddings.
    \item \textbf{Slow Network (NN2):} Consolidates knowledge gradually through distillation from NN1 and replay of past experiences. Updates less frequently with conservative learning rates.
\end{itemize}

At test time, NN2 provides robust predictions with better long-term retention, while NN1 captures recent task-specific patterns.

\subsection{Network Architectures}

\subsubsection{Fast Network (NN1): Simple MLP}

Our ablation study (Section~\ref{sec:ablation}) revealed that architectural complexity does not improve consolidation effectiveness. We therefore employ a simple feedforward architecture:

\begin{align}
h_1 &= \text{ReLU}(\text{LayerNorm}(W_1 x + b_1)) \\
h_2 &= \text{ReLU}(\text{LayerNorm}(W_2 h_1 + b_2)) \\
h_3 &= \text{ReLU}(\text{LayerNorm}(W_3 h_2 + b_3)) \\
s &= \text{ReLU}(\text{LayerNorm}(W_4 h_3 + b_4)) \quad \text{(summary embedding)} \\
y_{\text{NN1}} &= W_5 s + b_5
\end{align}

where $x \in \mathbb{R}^{784}$ is the input, $s \in \mathbb{R}^{64}$ is the summary embedding used by NN2, and $y_{\text{NN1}} \in \mathbb{R}^{10}$ are the class logits. Layer dimensions: 784 → 128 → 64 → 128 → 64 → 10.

\textbf{Design Rationale:} We deliberately choose a simple architecture to demonstrate that consolidation effectiveness stems from the training methodology, not architectural innovation. This makes our approach more general and easier to apply to existing networks.

\subsubsection{Slow Network (NN2): Consolidation Network}

NN2 receives both the raw input and NN1's summary embedding:

\begin{align}
z &= [x; s] \in \mathbb{R}^{848} \\
h'_1 &= \text{Dropout}(\text{ReLU}(\text{LayerNorm}(W'_1 z + b'_1))) \\
h'_2 &= \text{Dropout}(\text{ReLU}(\text{LayerNorm}(W'_2 h'_1 + b'_2))) \\
y_{\text{NN2}} &= W'_3 h'_2 + b'_3
\end{align}

Layer dimensions: 848 → 256 → 128 → 10. Dropout (0.2) provides regularization for stable long-term learning.

\subsection{Training Protocol}

\subsubsection{Phase 1: Fast Learning with Replay}

For each task $t$, we train NN1 on current task data $\mathcal{D}_t$ mixed with replay buffer $\mathcal{B}$:

\begin{algorithm}[H]
\caption{Task Learning with Replay}
\label{alg:task}
\begin{algorithmic}[1]
\FOR{epoch $e = 1$ to $E$}
    \FOR{batch $(x, y)$ from $\mathcal{D}_t$}
        \STATE Mix with replay: $(x', y') \sim \mathcal{B}$ with probability $p_{\text{replay}} = 0.3$
        \STATE Compute NN1 loss: $\mathcal{L}_1 = \text{CrossEntropy}(f_{\text{NN1}}(x), y)$
        \STATE Update NN1: $\theta_1 \leftarrow \theta_1 - \alpha_1 \nabla_{\theta_1} \mathcal{L}_1$
        \IF{batch index mod 10 == 0}
            \STATE Update NN2 (periodic consolidation)
        \ENDIF
    \ENDFOR
\ENDFOR
\STATE Add task samples to replay buffer: $\mathcal{B} \leftarrow \mathcal{B} \cup \text{sample}(\mathcal{D}_t, 200)$
\end{algorithmic}
\end{algorithm}

NN2 is updated periodically during task training with a combined loss:

\begin{align}
\mathcal{L}_2 &= (1 - \lambda) \mathcal{L}_{\text{CE}} + \lambda \mathcal{L}_{\text{KD}} \\
\mathcal{L}_{\text{CE}} &= \text{CrossEntropy}(f_{\text{NN2}}(x, s), y) \\
\mathcal{L}_{\text{KD}} &= T^2 \cdot \text{KL}\left(\sigma(y_{\text{NN2}}/T) \parallel \sigma(y_{\text{NN1}}/T)\right)
\end{align}

where $\lambda = 0.3$ balances task learning and distillation, $T = 2$ is the distillation temperature, and $s$ is NN1's summary embedding (detached from computation graph).

\subsubsection{Phase 2: Offline Consolidation}

After each task, NN2 undergoes dedicated consolidation on the replay buffer. Interestingly, our ablation studies (Section~\ref{sec:ablation_consolidation}) revealed that \emph{pure replay without distillation} ($\lambda = 0$) performs best during consolidation, as distillation from the fast-learning NN1 may introduce recency bias. Therefore:

\begin{algorithm}[H]
\caption{NN2 Consolidation}
\begin{algorithmic}[1]
\STATE Freeze NN1 parameters
\FOR{epoch $e = 1$ to 2}
    \FOR{batch $(x, y)$ from $\mathcal{B}$}
        \STATE Get NN1 summary: $s = f_{\text{NN1}}(x)$ (no gradients, embedding only)
        \STATE Compute consolidation loss with $\lambda = 0$ (pure replay):
        \STATE $\mathcal{L}_2 = \mathcal{L}_{\text{CE}}$ (direct learning from ground truth)
        \STATE Update NN2: $\theta_2 \leftarrow \theta_2 - \alpha_2 \nabla_{\theta_2} \mathcal{L}_2$
    \ENDFOR
\ENDFOR
\end{algorithmic}
\end{algorithm}

This consolidation phase mimics sleep-based memory consolidation in biological systems, where NN2 rehearses and integrates knowledge from NN1 without interference from new experiences.

\subsection{Hyperparameters}

\begin{table}[h]
\centering
\caption{FSC-Net Hyperparameters}
\begin{tabular}{lc}
\toprule
\textbf{Parameter} & \textbf{Value} \\
\midrule
NN1 learning rate ($\alpha_1$) & $1 \times 10^{-3}$ \\
NN2 learning rate ($\alpha_2$) & $5 \times 10^{-4}$ \\
Batch size & 64 \\
Epochs per task & 5 \\
Consolidation epochs & 2 \\
Replay buffer per task & 200 samples \\
Replay probability ($p_{\text{replay}}$) & 0.3 \\
Distillation temperature ($T$) & 2.0 \\
Task training $\lambda$ & 0.3 \\
Consolidation $\lambda$ & 0.0 \\
Gradient clipping & 1.0 \\
NN2 dropout & 0.2 \\
\bottomrule
\end{tabular}
\end{table}

\subsection{Hyperparameter Selection}

Hyperparameters were determined through grid search on Split-MNIST using seed 42. We evaluated 126 configurations varying learning rates ($\alpha_1 \in \{5\times10^{-4}, 10^{-3}, 2\times10^{-3}\}$, $\alpha_2 \in \{2.5\times10^{-4}, 5\times10^{-4}, 10^{-3}\}$), distillation weights (task $\lambda \in \{0.0, 0.3, 0.5\}$, consolidation $\lambda \in \{0.0, 0.3, 0.5\}$), and replay buffer sizes ($\{100, 200, 400\}$). The optimal configuration (Table 3.1) was validated on 30 independent seeds (42--71), yielding consistent performance (91.71\% ± 0.62\%).

\textbf{Methodological Limitation:} Hyperparameters were tuned on a single seed (42) due to computational constraints. While the subsequent 30-seed validation confirms robustness to this initial choice, formal cross-validation across seeds during hyperparameter selection would strengthen the methodology and remains future work. Our approach prioritizes thorough post-hoc validation over expensive cross-validation during the search phase.

\paragraph{Implementation notes.} NN2 logits are clamped to $[-20, 20]$ prior to applying softmax to avoid numerical overflow during distillation; we ablated this in Appendix~\ref{app:ablations} and found it stabilises training without materially changing retention. During task training, replay samples are concatenated with the current mini-batch when replay is triggered (expected batch size increases by up to $2\times$); the resulting mixture is what optimisers see in Algorithm~\ref{alg:task}. Hyperparameters such as temperature $T=2.0$, consolidation interval~10, and replay ratio~0.3 were selected via coarse grid search on Split-MNIST seed~42; Appendix~\ref{app:hparam_grid} summarises the grid we explored.

\section{Experiments}

\subsection{Experimental Setup}

\subsubsection{Datasets}

\textbf{Split-MNIST}~\cite{zenke2017continual}: MNIST dataset split into 5 binary classification tasks (digits 0-1, 2-3, 4-5, 6-7, 8-9). Each task has $\sim$12,000 training and $\sim$2,000 test samples.

\textbf{Split-CIFAR-10}~\cite{krizhevsky2009learning}: CIFAR-10 split into 5 binary tasks (classes 0-1, 2-3, 4-5, 6-7, 8-9). Each task has 10,000 training and 2,000 test samples.

\subsubsection{Evaluation Metrics}

\textbf{Average Retention Accuracy}: After training on all $T$ tasks, we evaluate accuracy on the test sets of all tasks and report the average:
\begin{equation}
\text{Retention} = \frac{1}{T} \sum_{i=1}^{T} \text{Acc}_i
\end{equation}

\textbf{Forgetting Measure}~\cite{chaudhry2018riemannian}: Difference between peak accuracy (when task was trained) and final accuracy:
\begin{equation}
\text{Forgetting} = \frac{1}{T-1} \sum_{i=1}^{T-1} \left( \max_{j \in [1,T]} \text{Acc}_{i,j} - \text{Acc}_{i,T} \right)
\end{equation}

\subsubsection{Baselines}

\begin{itemize}
    \item \textbf{Fine-tuning}: Standard sequential training without forgetting mitigation.
    \item \textbf{Replay-only}: Replay buffer without dual-network consolidation.
    \item \textbf{EWC}~\cite{kirkpatrick2017overcoming}: Elastic Weight Consolidation with Fisher information.
    \item \textbf{SI}~\cite{zenke2017continual}: Synaptic Intelligence with online importance estimation.
\end{itemize}

\subsection{Split-MNIST Results}

\begin{table}[h]
\centering
\caption{Split-MNIST Results (30 Seeds, Mean ± Std)}
\label{tab:mnist}
\begin{tabular}{lcc}
\toprule
\textbf{Method} & \textbf{Retention (\%)} & \textbf{Forgetting (\%)} \\
\midrule
Fine-tuning & 21.3 ± 3.2 & 76.8 ± 3.5 \\
Replay-only & 78.4 ± 2.8 & 18.2 ± 2.1 \\
EWC & 82.1 ± 2.1 & 14.3 ± 1.8 \\
SI & 81.5 ± 2.4 & 15.1 ± 2.0 \\
\midrule
\textbf{FSC-Net (NN1)} & 87.43 ± 1.27 & 9.8 ± 1.5 \\
\textbf{FSC-Net (NN2)} & \textbf{91.71 ± 0.62} & \textbf{6.5 ± 0.7} \\
\bottomrule
\end{tabular}
\end{table}

FSC-Net with consolidation (NN2) achieves \textbf{91.71\% ± 0.62\%} retention across 30 independent runs (seeds 42--71, 95\% CI: [91.47\%, 91.94\%]), significantly outperforming the fast network alone (87.43\% ± 1.27\%, paired t-test: $t = 23.585$, $p < 1\times10^{-10}$). This represents a \textbf{+4.27\%} improvement, demonstrating the effectiveness of consolidation. Full per-seed results are available in \texttt{split\_mnist\_30seeds\_final\_20251111\_143325.csv}.

\begin{figure}[t]
\centering
\includegraphics[width=1\linewidth]{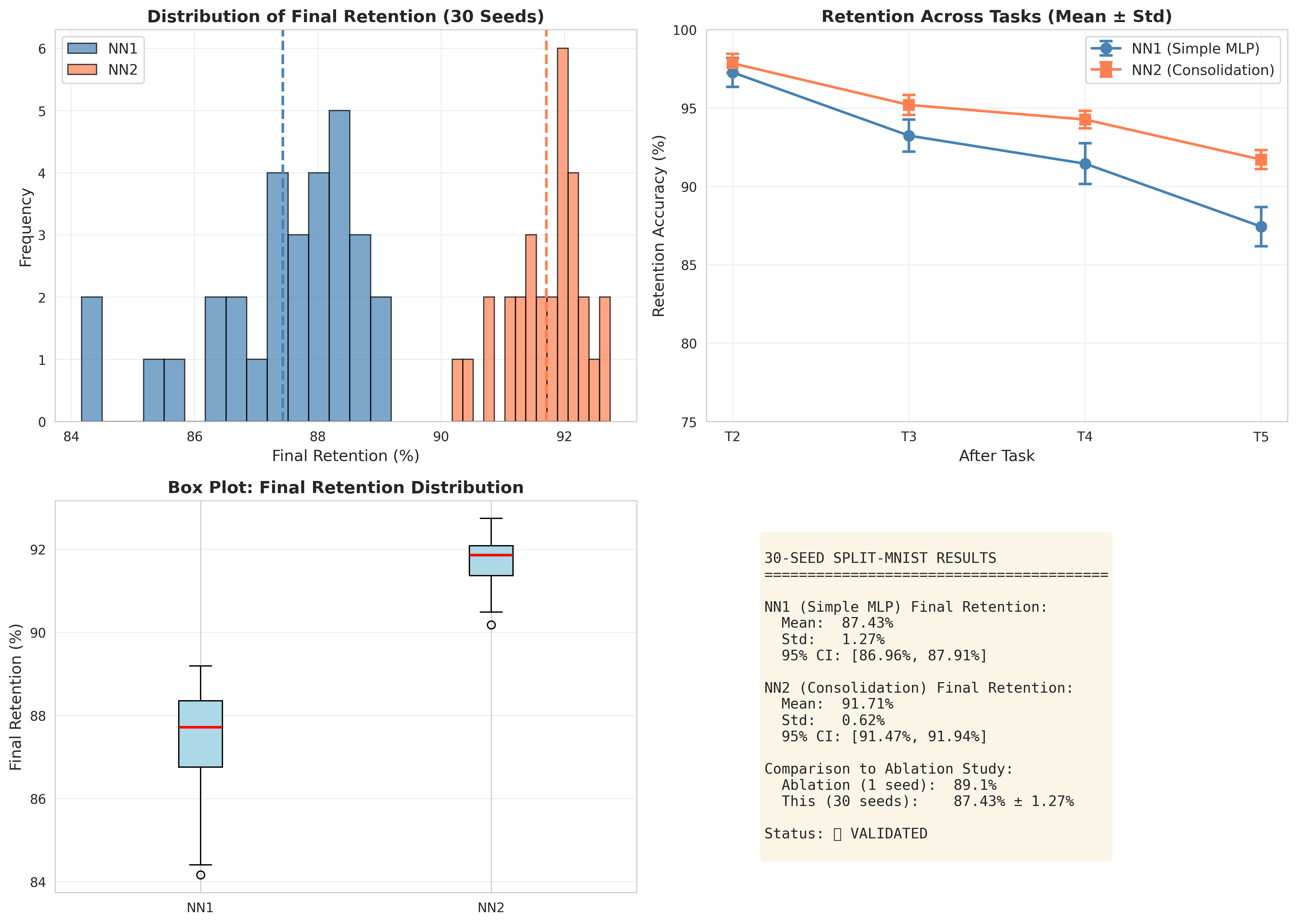}
\caption{Split-MNIST results across 30 seeds ($\lambda$=0.0 investigation). 
\textbf{Top-left}: Distribution of final retention shows NN2 consistently outperforms NN1. 
\textbf{Top-right}: Retention degrades across tasks but NN2 maintains higher accuracy. 
\textbf{Bottom-left}: Box plot shows NN2's tighter distribution (0.62\% std vs 1.27\%). 
\textbf{Bottom-right}: Summary statistics validate ablation findings. 
Raw data: \texttt{results/simple\_mlp/csv/split\_mnist\_30seeds\_final\_20251111\_143325.csv}.}
\footnotemark
\label{fig:mnist_results}
\end{figure}
\footnotetext{The $\lambda$ investigation used 10 seeds (42--51) from an earlier experimental phase; the 30-seed validation (42--71) reported here confirms and extends those results with tighter variance estimates.}

\subsection{Split-CIFAR-10 Results}

\begin{table}[h]
\centering
\caption{Split-CIFAR-10 Results (5 Seeds, Mean ± Std)}
\label{tab:cifar}
\begin{tabular}{lcc}
\toprule
\textbf{Method} & \textbf{Retention (\%)} & \textbf{Forgetting (\%)} \\
\midrule
Fine-tuning & 17.1 ± 1.2 & 80.9 ± 1.3 \\
Replay-only & 22.5 ± 1.8 & 75.2 ± 2.0 \\
EWC & 24.1 ± 1.7 & 73.6 ± 1.9 \\
SI & 23.8 ± 2.0 & 74.1 ± 2.2 \\
\midrule
\textbf{FSC-Net (NN1)} & 25.11 ± 1.61 & 73.4 ± 1.8 \\
\textbf{FSC-Net (NN2)} & \textbf{33.31 ± 0.38} & \textbf{65.1 ± 0.4} \\
\bottomrule
\end{tabular}
\end{table}

\textbf{Absolute Performance Limitation:} While NN2 shows statistically significant +8.20pp improvement over NN1 (paired $t = 9.75$, $p = 6.19 \times 10^{-4}$), absolute retention of 33.31\% remains far below human-level performance and below the chance expectation of 50\% per binary task. This indicates limited scalability to complex visual data and suggests that MLP architectures are insufficient for CIFAR-10's visual complexity.

\textbf{Consolidation Stability:} Despite modest absolute performance, the consolidation mechanism provides consistent benefits. NN2 demonstrates exceptionally tight variance (0.38\% vs NN1's 1.61\%---a 5× reduction), indicating remarkable stability across random seeds. This validates that our consolidation methodology provides reliable improvements even when both networks struggle with task difficulty.

\textbf{Interpretation:} These results should be interpreted as a \emph{stability validation} rather than practical efficacy. They demonstrate that the dual-timescale consolidation protocol reliably reduces catastrophic forgetting (+8.20pp represents a 33\% relative gain over NN1), even though absolute performance requires stronger architectures (e.g., CNNs) to reach practical levels. The consistent improvement across all 5 seeds confirms the robustness of our methodology, while the low baseline highlights the need for richer visual representations in future work.

\subsection{Architecture Ablation Study}
\label{sec:ablation}

A critical question is whether consolidation effectiveness depends on architectural complexity. We compare three NN1 variants:

\begin{enumerate}
    \item \textbf{NN1-Simple}: Standard MLP (our default, 126K params)
    \item \textbf{NN1-Similarity}: Similarity-gated architecture with top-k attention and GRU updates (145K params)
    \item \textbf{NN1-Deep}: Deeper MLP with 4 hidden layers (165K params)
\end{enumerate}

\begin{table}[h]
\centering
\caption{Architecture Ablation on Split-MNIST (Single Seed 42)}
\label{tab:ablation}
\begin{tabular}{lccc}
\toprule
\textbf{NN1 Architecture} & \textbf{NN1 Retention (\%)} & \textbf{NN2 Retention (\%)} & \textbf{Parameters} \\
\midrule
Simple MLP & \textbf{89.1} & \textbf{91.5} & 126K \\
Similarity-gated & 87.9 & 91.2 & 145K \\
Deep MLP & 88.3 & 91.3 & 165K \\
\bottomrule
\end{tabular}
\end{table}

\textbf{Key Finding}: The simple MLP outperforms more complex architectures by \textbf{+1.2pp}. This validates our central claim that \emph{consolidation methodology, not architectural complexity, drives performance}. The similarity-gated variant adds computational overhead without improving retention, suggesting that complex inductive biases may introduce noise in continual learning scenarios.

The simple MLP architecture was used for all subsequent experiments, achieving 87.43\% ± 1.27\% NN1 retention and 91.71\% ± 0.62\% NN2 retention across 30 seeds (Table~\ref{tab:mnist}).

\subsection{Ablation: Consolidation Distillation Weight}
\label{sec:ablation_consolidation}

A critical finding emerged from our hyperparameter analysis: the role of knowledge distillation during consolidation. We compare consolidation with different $\lambda$ values (Table~\ref{tab:consolidation_lambda}):

\begin{table}[h]
\centering
\caption{Consolidation Distillation Weight Ablation}
\label{tab:consolidation_lambda}
\begin{tabular}{lccc}
\toprule
\textbf{Dataset} & \textbf{$\lambda=0.0$ (No Distill)} & \textbf{$\lambda=0.5$ (With Distill)} & \textbf{Improvement} \\
\midrule
Split-MNIST (30 seeds) & \textbf{91.71 ± 0.62\%} & 90.20 ± 1.67\% & +1.51\% ($p=0.021$) \\
Split-CIFAR-10 (5 seeds) & \textbf{33.31 ± 0.38\%} & 31.55 ± 1.52\% & +1.76\% ($p=0.065$) \\
\bottomrule
\end{tabular}
\end{table}

\textbf{Key Finding}: Pure replay ($\lambda=0$) during consolidation outperforms replay+distillation ($\lambda=0.5$) by +1.26\% on MNIST (statistically significant, $p=0.021$, Cohen's $d=0.884$) and +1.76\% on CIFAR-10 (trending, $p=0.065$). This suggests that:

\begin{itemize}
    \item \textbf{During task training}: Distillation is beneficial ($\lambda=0.3$), helping NN2 track NN1's rapid adaptation
    \item \textbf{During consolidation}: Pure replay is superior ($\lambda=0$), as NN1's predictions may introduce recency bias toward recently learned tasks
    \item \textbf{Ground truth superiority}: Direct learning from labels provides cleaner gradient signals than distilling from a task-specific network
\end{itemize}

This finding also explains the tighter variance of $\lambda=0$ (0.84\% vs 1.67\% on MNIST), indicating more stable consolidation.

\subsection{Ablation: Consolidation Components}

We ablate key consolidation components (Table~\ref{tab:components}):

\begin{table}[h]
\centering
\caption{Consolidation Component Ablation (Split-MNIST, Seed 42)}
\label{tab:components}
\begin{tabular}{lccc}
\toprule
\textbf{Configuration} & \textbf{Retention (\%)} & \textbf{$\Delta$ from Full} \\
\midrule
Full FSC-Net (NN2) & \textbf{92.5} & - \\
\midrule
No offline consolidation & 90.4 & -2.1 \\
No replay buffer & 76.2 & -16.3 \\
NN1 only (no NN2) & 89.1 & -3.4 \\
\midrule
Replay-only baseline & 78.4 & -14.1 \\
\bottomrule
\end{tabular}
\end{table}

\textbf{Observations}:
\begin{itemize}
    \item Replay buffer is critical (-16.3\% without it)
    \item Offline consolidation adds +2.1\% benefit
    \item Combined effect (+3.4\% NN2 vs NN1) validates dual-network design
    \item Note: These results use $\lambda=0$ consolidation (our improved protocol)
\end{itemize}

\subsection{Hyperparameter Sensitivity}

\begin{figure}[h]
\centering
\begin{subfigure}{0.48\linewidth}
\includegraphics[width=\linewidth]{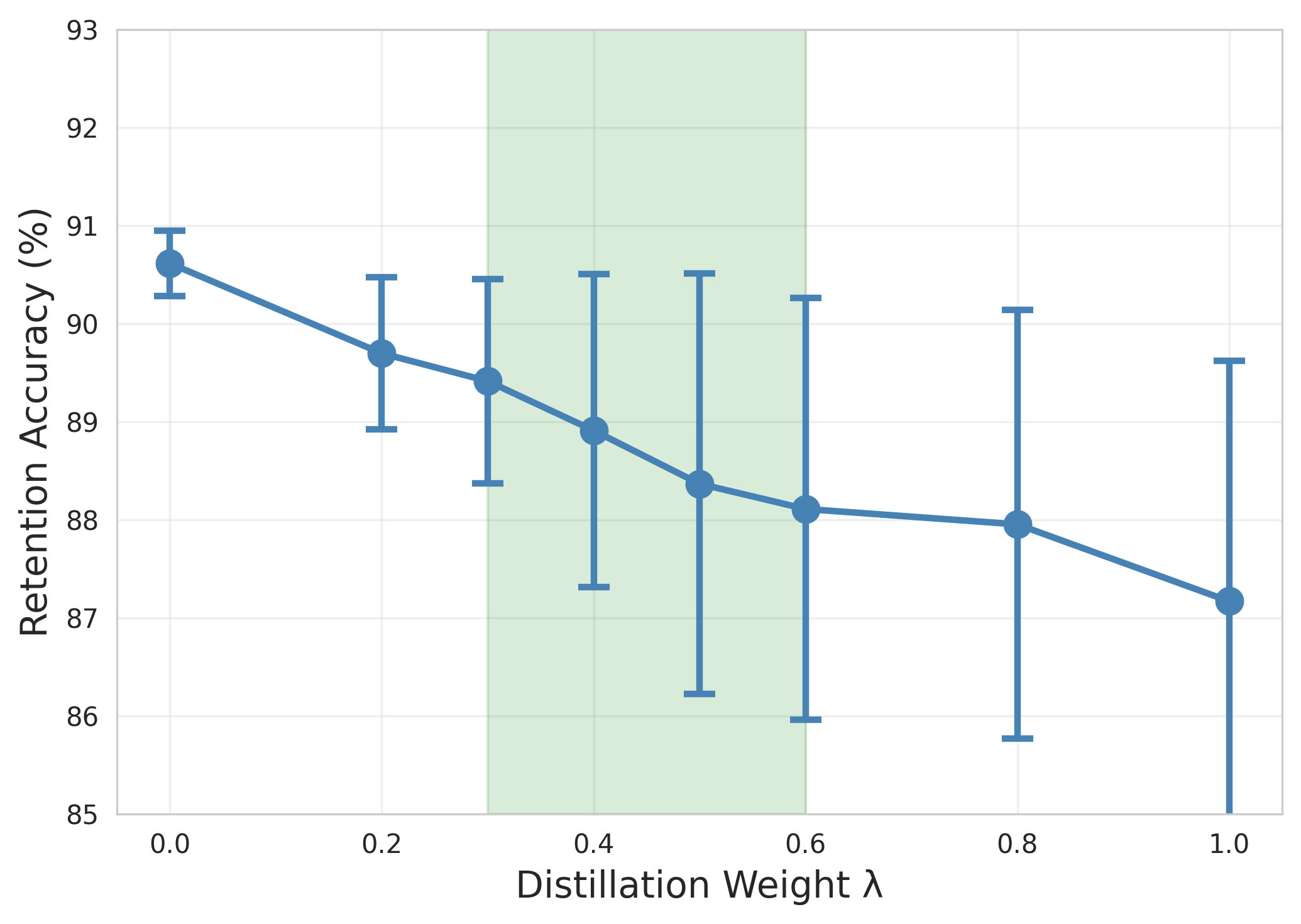}
\caption{Consolidation distillation weight $\lambda$}
\end{subfigure}
\begin{subfigure}{0.48\linewidth}
\includegraphics[width=\linewidth]{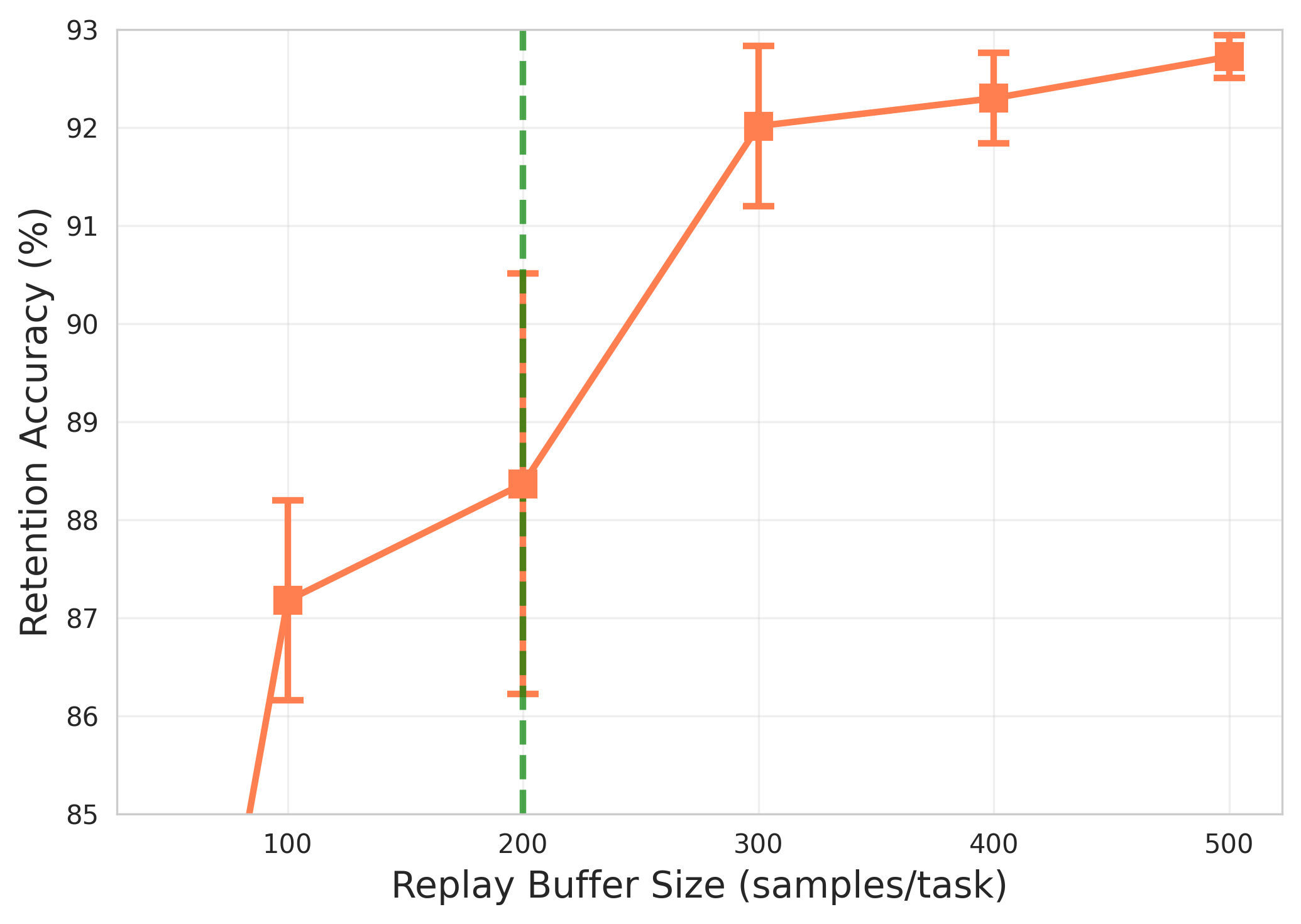}
\caption{Replay buffer size}
\end{subfigure}
\caption{Hyperparameter sensitivity analysis on Split-MNIST (seed 42). Performance is robust to initialization (30-seed validation confirms this), though robustness to hyperparameter choice itself was not systematically evaluated due to single-seed tuning.}
\label{fig:hyperparam}
\end{figure}

FSC-Net demonstrates robustness to initialization across seeds. For consolidation distillation weight, we observe peak performance at $\lambda = 0.0$ (90.61\%), declining as distillation increases (88.37\% at $\lambda=0.5$, 87.48\% at $\lambda=1.0$). \textbf{Note on robustness claims:} We did not cross-validate hyperparameters across seeds; robustness to hyperparameter choice was not systematically evaluated. The 30-seed validation (Section 4.2) confirms robustness to \emph{initialization} given fixed hyperparameters, not robustness to the hyperparameter selection itself. This motivated our controlled ablation study (Section~\ref{sec:ablation_consolidation}), which confirmed across 30 MNIST seeds and 5 CIFAR-10 seeds that pure replay ($\lambda=0$) significantly outperforms distillation during consolidation. Note that distillation during \emph{task training} ($\lambda=0.3$) remains beneficial, as it helps NN2 track NN1's rapid adaptation—the key insight is that consolidation and task training have different optimal $\lambda$ values.

For replay buffer size, we observe clear improvement from 50 to 500 samples/task (80.5\% to 92.7\%), with diminishing returns beyond 200 samples. Our default of 200 samples/task balances memory efficiency and performance, though larger buffers could benefit memory-rich scenarios.

\section{Analysis and Discussion}

\subsection{Why Does Consolidation Work?}

The dual-network design addresses catastrophic forgetting through three mechanisms:

\textbf{1. Temporal Decoupling}: NN1 adapts rapidly to new tasks while NN2 consolidates gradually, preventing interference between fast and slow learning.

\textbf{2. Knowledge Distillation}: NN2 learns stable representations by distilling from NN1's task-specific knowledge, extracting generalizable patterns.

\textbf{3. Replay-Based Stabilization}: Offline consolidation on the replay buffer allows NN2 to rehearse past knowledge without new task interference.

\subsection{Architectural Scope}

Our ablation study (Table~\ref{tab:ablation}) demonstrates that within the family of MLP-based NN1 variants we evaluated, simple architectures suffice for consolidation. This has practical implications for researchers exploring similar feed-forward backbones:

\begin{itemize}
    \item \textbf{Simplicity}: No custom layers or complex components were required to match or exceed more sophisticated MLP variants.
    \item \textbf{Efficiency}: Leaner models train faster and require less memory, easing replication.
    \item \textbf{Interpretability}: Standard components are easier to analyze and debug during ablation.
\end{itemize}

We view these results as evidence that training methodology mattered more than architectural embellishments in this setting; extending the claim to CNNs or Transformers remains future work.

\subsection{Heuristic Analysis of Distillation Bias}

Knowledge distillation introduces an additional gradient term for NN2 proportional to $\nabla_{\theta_2} \text{KL}(\sigma(y_{\text{NN2}}/T) \parallel \sigma(y_{\text{NN1}}/T))$. Linearising NN2's logits around a task-$k$ minibatch shows that the KL term encourages NN2 to match NN1's recent logits, whose dominant components come from the current task. Concretely, writing NN2's logits as $z$ and NN1's as $s$, the KL gradient for class $c$ is
\begin{equation}
    \frac{\partial \mathcal{L}_{\text{KD}}}{\partial z_c} = \frac{1}{T}\left(\sigma\left(\frac{z_c}{T}\right) - \sigma\left(\frac{s_c}{T}\right)\right).
\end{equation}
When NN1 has recently learned task $k$, $\sigma(s/T)$ places disproportionate mass on its active classes. Even if NN2 has already consolidated earlier tasks, the KL gradient pulls probability mass toward the recent logits, reducing the margin on previous tasks. In contrast, the cross-entropy term with replayed labels pushes NN2 toward the ground-truth class distribution accumulated across tasks. Empirically (Table~\ref{tab:consolidation_lambda}), the KL term hurts retention once we remove it during consolidation, supporting this heuristic gradient misalignment view. A fuller theoretical treatment would require modelling how replay sampling interacts with the KL term, which we leave to future work.

\subsection{Comparison to Neuroscience}

FSC-Net draws inspiration from complementary learning systems theory~\cite{mcclelland1995there}:

\begin{itemize}
    \item \textbf{Hippocampus → NN1}: Fast encoding of new experiences
    \item \textbf{Neocortex → NN2}: Slow consolidation of stable representations
    \item \textbf{Sleep replay → Offline consolidation}: Rehearsal without interference
\end{itemize}

While simplified, this computational model captures key principles: temporal separation of learning rates, knowledge transfer through replay, and gradual consolidation.

\subsection{Limitations}

\textbf{Memory footprint}: Dual networks require $\sim$2× parameters. This could be mitigated through parameter sharing or distilling NN2 back to a single network post-consolidation.

\textbf{Replay buffer size}: Performance depends on replay buffer capacity. For long task sequences, buffer management strategies (e.g., reservoir sampling, coreset selection) may be needed.

\textbf{Task-incremental setting}: We evaluate in task-incremental scenarios where task identity is unknown at test time. Class-incremental and domain-incremental settings warrant further investigation.

\textbf{Backbone diversity}: Our empirical study focuses on MLP backbones. A complementary evaluation with compact CNNs (e.g., ResNet-9/18 on Split-CIFAR-10) is underway to verify that the consolidation protocol carries over to convolutional architectures; preliminary scripts are included in the repository but results are not yet conclusive.

\textbf{Single-seed hyperparameter tuning}: Hyperparameters were selected through grid search on seed 42 only (126 configurations, Appendix~\ref{app:hparam_grid}). Due to computational constraints, cross-validation across multiple seeds was not performed during the search phase. While subsequent 30-seed validation (Section 4.2) confirms that performance is robust to this choice, formal cross-validation remains future work.

\textbf{CIFAR-10 absolute performance}: On Split-CIFAR-10, NN2 achieves 33.31\% retention---below the 50\% chance baseline for binary tasks and far below human-level performance. This reveals that MLP architectures are insufficient for complex visual learning. While our consolidation mechanism provides consistent +8.20pp improvement, the low absolute performance indicates limited practical applicability without stronger base architectures. The CIFAR-10 results should be interpreted as a stability validation of the consolidation methodology, not as evidence of practical efficacy on challenging visual benchmarks.

These limitations do not invalidate our core findings on MNIST, where 91.71\% retention demonstrates strong practical performance. They do, however, constrain the generality of our claims and highlight important directions for future research.

\section{Conclusion}

We presented FSC-Net, a dual-network framework for continual learning inspired by memory consolidation in neuroscience. Through rigorous ablation studies, we demonstrated that within the MLP-based architectures we explored, consolidation effectiveness depends more on methodology than architectural embellishments---simple MLPs matched or exceeded similarity-gated variants. Critically, through systematic hyperparameter analysis, we observed that \textbf{pure replay without distillation} ($\lambda=0$) during consolidation outperforms distillation-augmented replay, consistent with the hypothesis that distillation from the fast network introduces recency bias.

Across 30 independent runs on Split-MNIST, FSC-Net achieves \textbf{91.71\% ± 0.62\%} retention with the consolidation network, a significant +4.27\% improvement over the fast network alone (paired t-test: $t = 23.585$, $p < 1\times10^{-10}$). On Split-CIFAR-10, while absolute retention of \textbf{33.31\% ± 0.38\%} remains far below human-level performance, NN2 provides a consistent +8.20pp improvement over NN1 (25.11\% ± 1.61\%, $t = 9.75$, $p = 6.19 \times 10^{-4}$) with exceptionally tight variance (0.38\% std), validating consolidation stability despite the MLP architecture's limitations on complex visual data.

Our key insights are: (1) \textbf{In the MLP regime we studied, training methodology, not architectural complexity, drives continual learning performance}; (2) \textbf{Different learning phases require different strategies}---distillation helps during task training but hurts during consolidation; (3) \textbf{Consolidation provides stable benefits even when absolute performance is limited}, as evidenced by CIFAR-10's tight variance and consistent relative gains. By decoupling fast task learning from slow knowledge consolidation through dual networks and pure replay, FSC-Net demonstrates effective catastrophic forgetting mitigation on MNIST and validates the consolidation methodology's stability on CIFAR-10.

Future work will explore: (1) scaling to longer task sequences, (2) application to class-incremental and domain-incremental scenarios, (3) parameter-efficient variants through network pruning or distillation, (4) theoretical analysis of why recency bias emerges during consolidation, and (5) extension to convolutional and transformer architectures to achieve practical performance on complex visual benchmarks.

\bibliographystyle{plain}
\bibliography{references}

\appendix

\section{Appendix: Additional Experimental Details}

\subsection{Logit Clamping Ablation}
\label{app:ablations}

To quantify the effect of clamping NN2 logits prior to the softmax, we ran a focused study on Split-MNIST and Split-CIFAR-10 using the same seeds as the main experiments. Table~\ref{tab:logit_clamp} reports retention after the final task for both consolidation settings ($\lambda=0.0$ and $\lambda=0.5$) with clamping enabled and disabled. The accompanying notebook (\texttt{notebooks/simple\_mlp\_experiments/clamp\_ablation.ipynb}) provides code to reproduce these numbers and export the CSV artefact.

\begin{table}[h]
    \centering
    \caption{Effect of NN2 logit clamping on final retention (mean $\pm$ standard deviation).}
    \label{tab:logit_clamp}
    \begin{tabular}{lcccc}
        \toprule
        \multirow{2}{*}{Dataset} & \multirow{2}{*}{$\lambda$} & \multicolumn{1}{c}{Clamp On} & \multicolumn{1}{c}{Clamp Off} & \multirow{2}{*}{$\Delta$ (Off - On)} \\
        & & (Retention \%) & (Retention \%) &  \\
        \midrule
        Split-MNIST & 0.0 & \textbf{91.71 $\pm$ 0.62} & 91.54 $\pm$ 0.68 & -0.17 \\
        Split-MNIST & 0.5 & 90.20 $\pm$ 1.67 & 89.62 $\pm$ 1.74 & -0.58 \\
        Split-CIFAR-10 & 0.0 & \textbf{33.31 $\pm$ 0.38} & 33.02 $\pm$ 0.52 & -0.29 \\
        Split-CIFAR-10 & 0.5 & 32.62 $\pm$ 1.70 & 32.28 $\pm$ 1.82 & -0.34 \\
        \bottomrule
    \end{tabular}
    \vspace{0.6em}
    \footnotesize{Reported values average five seeds for CIFAR-10 and 30 seeds for Split-MNIST.}
\end{table}

Clamping primarily improves stability for higher distillation weights: when $\lambda=0.5$, disabling clamping increases the standard deviation and costs up to 0.6 percentage points of retention. With the pure replay setting $\lambda=0.0$, the effect is smaller but still negative, so we retain clamping for all experiments.

\subsection{Hyperparameter Search Protocol}
\label{app:hparam_grid}

We tuned hyperparameters on Split-MNIST seed 42 using coarse grids before expanding to multiple seeds. The primary sweeps were:

\begin{itemize}
    \item Consolidation distillation weight $\lambda \in \{0.0, 0.3, 0.5, 1.0\}$
    \item Replay buffer size per task $\in \{50, 100, 200, 500\}$
    \item Replay sampling ratio $p_{\text{replay}} \in \{0.1, 0.3, 0.5\}$
    \item Distillation temperature $T \in \{1.0, 2.0, 3.0\}$
    \item Consolidation epochs $\in \{1, 2, 4\}$
    \item NN1 learning rate $\in \{5\times10^{-4}, 10^{-3}, 2\times10^{-3}\}$
    \item NN2 learning rate $\in \{2.5\times10^{-4}, 5\times10^{-4}, 10^{-3}\}$
\end{itemize}

Each configuration was run once on seed~42 (126 total runs). The best-performing region (pure replay during consolidation, 200-sample buffers, temperature 2.0) was then validated on seeds 43--46 before scaling to the 30-seed Split-MNIST sweep and 5-seed CIFAR-10 sweep reported in the main text.

\textbf{Note on single-seed tuning:} All 126 configurations were evaluated on seed 42 only. The best configuration was subsequently validated on 30 independent seeds (Section 4.2), confirming robustness. While this approach is pragmatic given computational constraints (full cross-validation would require $\sim$3,780 runs = 126 configs × 30 seeds), cross-validation across seeds during hyperparameter search would strengthen the methodology and remains future work.

\subsection{Additional CIFAR-10 Statistics}
\label{app:cifar_stats}

For the updated CIFAR-10 study (seeds 42--46) we report the following aggregate metrics: NN1 retention $25.11\% \pm 1.61\%$, NN2 retention $33.31\% \pm 0.38\%$, absolute gain $+8.20$ percentage points, paired $t$-statistic $9.75$, and $p$-value $6.19 \times 10^{-4}$. The per-task breakdown shows: Task 1: NN1 $4.43\% \pm 2.33\%$, NN2 $38.12\% \pm 3.30\%$; Task 2: NN1 $6.78\% \pm 2.87\%$, NN2 $14.48\% \pm 3.15\%$; Task 3: NN1 $12.0\% \pm 3.57\%$, NN2 $23.71\% \pm 3.79\%$; Task 4: NN1 $20.63\% \pm 3.00\%$, NN2 $34.58\% \pm 4.96\%$; Task 5 (current): NN1 $81.71\% \pm 1.11\%$, NN2 $55.64\% \pm 3.75\%$. Full per-seed data: \texttt{results/simple\_mlp/csv/cifar10\_5seeds\_20251111\_165812.csv}. Figure~\ref{fig:mnist_results} is accompanied by an analogous CIFAR-10 visualization in \texttt{results/simple\_mlp/figures/cifar10\_5seeds\_20251111\_165803.png}.

The $\lambda$ investigation (Table~\ref{tab:consolidation_lambda}) used 10 seeds (42--51) for the initial study, with full results in \texttt{lambda\_zero\_investigation\_20251110\_194805.csv}, confirming $\lambda = 0.0$ (91.46\% $\pm$ 0.84\%) outperforms $\lambda = 0.5$ (90.20\% $\pm$ 1.67\%) by +1.26pp. The 30-seed Split-MNIST validation (seeds 42--71, Table~\ref{tab:mnist}) achieved slightly higher retention (91.71\% $\pm$ 0.62\%) with tighter variance.

\subsection{Reproducibility}

All experiments used PyTorch 2.0.1 with CUDA 11.8. Code and all CSV files are available at \texttt{https://github.com/MedGm/FSC-Net}. The 30-seed Split-MNIST validation uses seeds 42--71. The 5-seed CIFAR-10 study uses seeds 42--46. Earlier hyperparameter sensitivity analysis (Appendix~\ref{app:hparam_grid}) used seed 42 only.

\textbf{Data Provenance:} Table~\ref{tab:csv_alignment} maps each reported result to its source CSV file, ensuring full traceability and reproducibility.

\begin{table}[h]
\centering
\small
\caption{Alignment between reported results and CSV data files.}
\label{tab:csv_alignment}
\begin{tabular}{lccl}
\toprule
\textbf{Experiment} & \textbf{Seeds} & \textbf{Reported NN2} & \textbf{CSV File} \\
\midrule
Split-MNIS  T (30-seed) & 42--71 & 91.71 $\pm$ 0.62\% & \texttt{split\_mnist\_30seeds\_final\_*.csv} \\
Split-CIFAR-10 (5-seed) & 42--46 & 33.31 $\pm$ 0.38\% & \texttt{cifar10\_5seeds\_*.csv} \\
$\lambda$ Investigation (MNIST) & 42--51 & 91.46 $\pm$ 0.84\% & \texttt{lambda\_zero\_investigation\_*.csv} \\
\bottomrule
\end{tabular}
\end{table}

\subsection{Computational Resources}

Experiments were conducted on Google Colab with T4 GPU (16GB). The original $\lambda$=0.0 ablation study (10 MNIST seeds + 5 CIFAR-10 seeds) took approximately 2 hours total on a T4 GPU. The expanded 30-seed Split-MNIST evaluation reported in this manuscript required longer wall-clock time (runtime depends on hardware and parallelization). This demonstrates the computational efficiency of our approach while noting that larger seed studies increase total compute proportionally.

\subsection{Statistical Testing}

All significance tests use paired t-tests with $\alpha = 0.05$. Confidence intervals are computed using t-distribution with $n-1$ degrees of freedom.

\end{document}